\documentclass[runningheads]{llncs}

\usepackage{eccv}

\usepackage{eccvabbrv}

\usepackage{graphicx}
\usepackage{wrapfig}
\usepackage{booktabs}
\usepackage{amsmath,amssymb}
\usepackage{multirow}
\usepackage{placeins}
\usepackage{float}
\usepackage{algorithm}
\usepackage{algpseudocode}
\usepackage{array}
\usepackage{xcolor}

\usepackage[accsupp]{axessibility}  

\usepackage[pagebackref,breaklinks,colorlinks,citecolor=eccvblue]{hyperref}

\usepackage{orcidlink}

\newcommand{\TileKey}[1]{\textcolor{blue!75}{#1}}
\newcommand{\EARKey}[1]{\textcolor{red!80}{#1}}

\newcommand{\QualLegendRow}{
\smallskip
\scriptsize
\noindent
\makebox[\linewidth][l]{%
\makebox[0pt][l]{\hspace*{0.090\linewidth}\makebox[0pt][c]{\raisebox{2.8ex}{\shortstack[t]{(a) Input\\(8-bit sRGB image)}}}}%
\makebox[0pt][l]{\hspace*{0.289\linewidth}\makebox[0pt][c]{\raisebox{2.8ex}{\shortstack[t]{(b) CAM~\cite{nam2022learning}\\(bpp: 0.8438)}}}}%
\makebox[0pt][l]{\hspace*{0.487\linewidth}\makebox[0pt][c]{\shortstack[t]{(c) Beyond-\\R2LCM~\cite{wang2024beyond}\\(bpp: 0.3763)}}}%
\makebox[0pt][l]{\hspace*{0.680\linewidth}\makebox[0pt][c]{\raisebox{2.8ex}{\shortstack[t]{(d) Ours\\(bpp: 0.3612)}}}}%
\makebox[0pt][l]{\hspace*{0.910\linewidth}\makebox[0pt][c]{\raisebox{2.8ex}{\shortstack[t]{(e) Raw image\\\phantom{(bpp: 0.3612)}}}}}%
}
\normalsize
}

\begin{document}

\title{MambaRaw: Selective State Space Modeling for Efficient 4K Raw Image Reconstruction} 

\titlerunning{MambaRaw}


\author{Peize Li\inst{1,2}\thanks{Equal Contribution. $^{\dag}$Corresponding Author.} \and
Fanhu Zeng\inst{1 \star} \and
Tongda Xu\inst{1}\and
Xingguo Xu\inst{3}\and
Xinjie Zhang\inst{4}\and
Xingtong Ge\inst{5}\and
Haotian Zhang\inst{6}\and
Yan Wang\inst{1}$^{\dagger}$}
\authorrunning{P. Li \etal}

\institute{Institute for AI Industry Research (AIR), Tsinghua University \and
King's College London \and 
Dalian University of Technology \and
Microsoft Research Asia \and
Hong Kong University of Science and Technology \and
School of Computer Science, Peking University}

\maketitle
\begin{abstract}
In-camera JPEG previews are ubiquitous in raw image formats and provide an sRGB reference at negligible storage cost. Although existing metadata-based reconstruction frameworks can exploit this side information when recovering raw images, their context models often become computationally expensive especially at high resolution, \eg, 4K raw image, given that attention mechanisms scale quadratically with feature maps, hindering its practical application.
To address these limitations, we propose \textbf{MambaRaw}, a JPEG-conditioned metadata-based raw image reconstruction framework that uses State Space Models (SSMs) to estimate entropy parameters efficiently. Our key contribution comprises a Spatial-Energy Coupled Context Modeling mechanism with two lightweight modules: (1) TileMambaBlock, which performs Mamba-style selective scanning only on information-dense tiles to improve the efficiency; and (2) Energy-Aware Refinement (EAR), an identity-initialized residual module that enhance feature representation to match the long-tail energy distribution of raw signals. Extensive experiments on three camera datasets (Sony, Olympus, Samsung) show consistent improvements over strong metadata-based baselines and set a new state of the art for JPEG-guided raw reconstruction with great efficiency. Notably, at low metadata bitrates, MambaRaw increases PSNR by 1.2--1.4 dB and reduces end-to-end coding latency by about 9\%. Code is released at \url{https://github.com/Peizeli1/MambaRaw}.
\end{abstract}

\section{Introduction}
Raw images preserve scene-referred radiance with high bit depth and dynamic range. They provide a high-fidelity basis for computational photography. However, storing and transmitting high-resolution raw captures requires substantial bandwidth. Standard codecs such as JPEG~\cite{wallace1991jpeg} and HEIF~\cite{warenkorb2015information} often perform poorly on raw data. The spatial and channel statistics of sRGB differ from those of raw signals, so sRGB-oriented encoders are not well matched. Learned image compression (LIC)~\cite{balle2018variational,minnen2018joint} improves efficiency by learning context models, but extending LIC to raw data remains challenging. Raw signals have uneven and camera-dependent channel distributions, which are not well captured by uniform context models.

\begin{wrapfigure}{R}{0.5\textwidth}
\centering
\includegraphics[width=\linewidth]{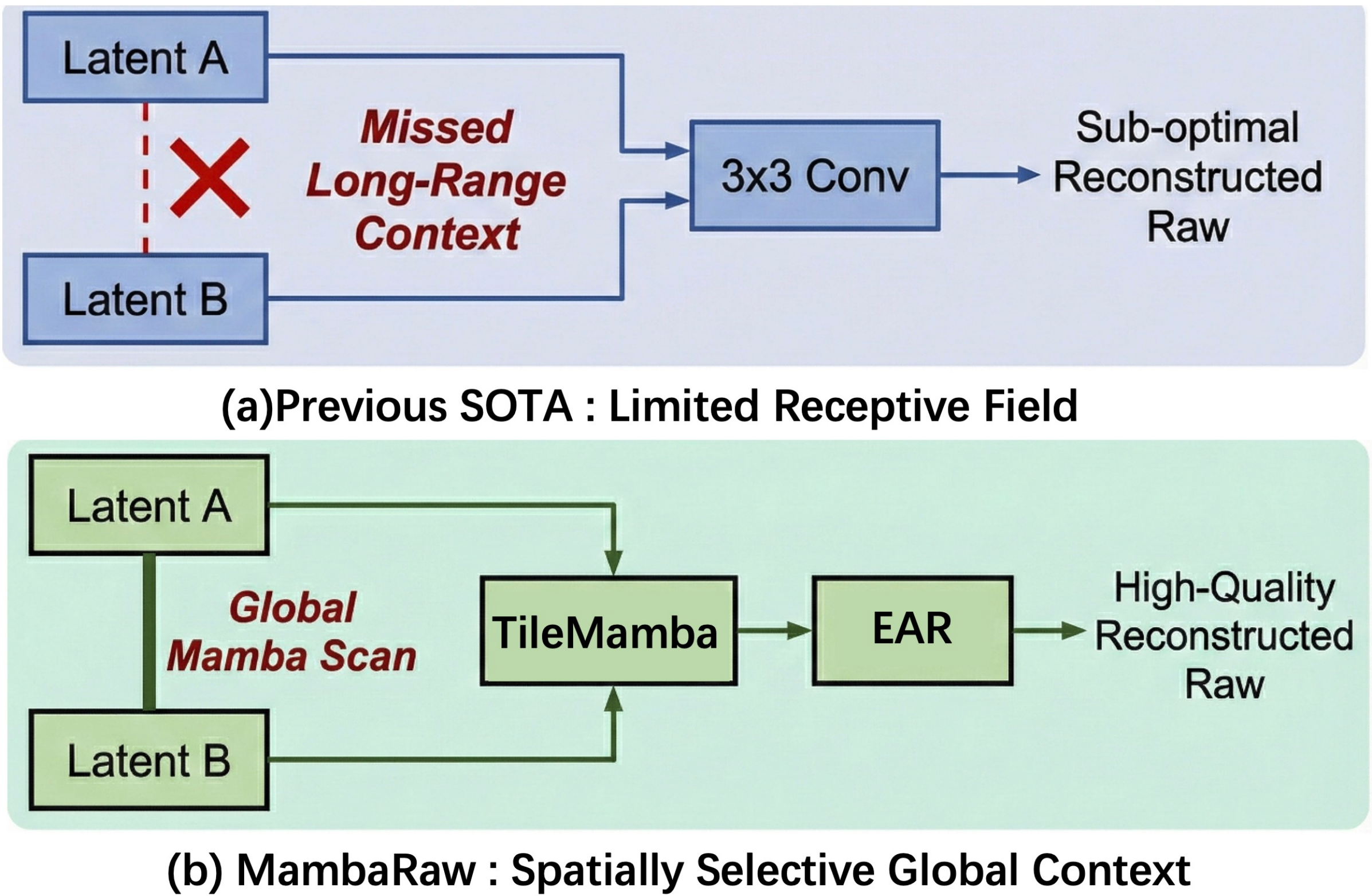}
\caption{Motivation and Comparison. (a) Convolution-based methods have limited receptive fields, which restricts long-range spatial modeling. (b) \textbf{MambaRaw} uses a spatial--energy coupled context model. It applies TileMambaBlock for selective scanning on information-dense tiles and uses EAR for energy-guided refinement.}
\label{fig:motivation}
\vskip -10pt
\end{wrapfigure}
Many raw formats also store an aligned in-camera JPEG preview. Recent metadata-based reconstruction frameworks, such as SAM~\cite{punnappurath2021spatially} and R2LCM~\cite{wang2023raw}, exploit this side information. They transmit a compact metadata bitstream and reconstruct raw signals with JPEG guidance. The main bottleneck is still context modeling at high resolution. As shown in Fig.~\ref{fig:motivation}\textcolor{red}{a}, convolutional context models have limited receptive fields and miss long-range spatial correlations. Self-attention is also costly because its compute and memory grow quadratically on 4K feature maps. In addition, prior methods apply these heavy context modules to all regions, which wastes computation. To this end, it is natural that raw reconstruction needs content-adaptive spatial modeling. Smooth regions require limited processing, while texture-rich regions benefit from accurate long-range reasoning. This motivates an efficient architecture that combines long-range sequence modeling with sparse computation.

Based on this observation, we propose \textbf{MambaRaw}, an efficient JPEG-conditioned metadata reconstruction framework that integrates state space models into entropy parameter estimation with Spatial-Energy Coupled Context Modeling. It includes two lightweight modules: (1) \textbf{TileMambaBlock}, which applies selective scanning to information-dense tiles only to reduce computation while preserving global context; (2) \textbf{Energy-Aware Refinement (EAR)}, an identity-initialized residual module that enhance features based on long-tail raw energy distribution. 

Extensive experiments on three diverse camera datasets (Sony, Olympus, and Samsung) demonstrate that MambaRaw establishes a new state of the art for JPEG-guided raw reconstruction. By effectively addressing the context modeling bottleneck, our approach achieves consistent rate--distortion improvements over strong metadata-based baselines. Notably, at competitive metadata bitrates, MambaRaw yields substantial PSNR gains of 1.2--1.4\,dB while simultaneously reducing the end-to-end coding latency by approximately 9\%, demonstrating a superior balance between high-fidelity reconstruction and practical computational efficiency.
Our contributions are as follows:
\begin{itemize}
    \item We propose a \textbf{spatial--energy coupled context modeling} paradigm for JPEG-guided raw reconstruction. It addresses high-resolution bottleneck by combining global spatial scanning with energy-guided entropy refinement.
    \item We introduce two lightweight modules: \textbf{TileMambaBlock} and \textbf{Energy-Aware Refinement (EAR)}, which enables content-adaptive selective scanning for efficient global modeling on information-dense tiles and enhances spatial entropy features for long-tail raw distributions.
    \item Extensive experiments show the effectiveness and efficiency of our method, with 1.2--1.4\,dB PSNR improvement over strong baselines and about 9\% lower end-to-end coding time.
\end{itemize}

\section{Related Work}
\subsection{Learned Image Compression}
\textbf{Learned image compression} advances rapidly since Ball\'e \emph{et al.}~\cite{balle2016end} introduce end-to-end neural image compression. Many follow-up studies improve entropy modeling with hyperpriors~\cite{balle2018variational}, joint autoregressive and hierarchical priors~\cite{minnen2018joint}, causal context prediction~\cite{guo2021causal}, and attention with mixture likelihoods~\cite{cheng2020learned,gao2021neural}. Other works focus on faster context designs, including checkerboard modeling~\cite{he2021checkerboard} and efficient convolutional entropy modeling~\cite{li2020efficient}. To better capture long-range dependencies, Transformer-based models are introduced for compression~\cite{zhu2022transformer,liu2023learned} and achieve improved rate--distortion performance. Uneven grouping and cross-channel context modeling~\cite{he2022elic,minnen2020channel,ma2021cross, qin2026freqsic} further improve the balance between spatial and channel aggregation. Recent methods also explore window attention and mixed Transformer--CNN designs~\cite{zou2022devil,liu2023learned}. Despite strong performance, these approaches can be expensive on high-resolution raw inputs because attention scales poorly and activation memory is large.
Beyond images, learned compression principles extend to video, including probabilistic video rescaling~\cite{tian2021self}, unsupervised video semantic compression~\cite{tian2023non,tian2024free,tian2025smc++}, and low-bitrate coding frameworks for video understanding~\cite{tian2024coding}, all highlighting the importance of effective entropy modeling and content-adaptive context designs.

\noindent \textbf{Efficient high-resolution processing}
is a common bottleneck in learned compression because context modeling operates on large feature maps. Prior works explore adaptive computation for efficient image restoration~\cite{zhou2024efficient}, C2SSM's cluster-centric scanning paradigm for ultra-high-definition image restoration~\cite{wu2026scan}, and saliency-driven bit allocation for perceptual compression~\cite{patel2021saliency}. Instead of applying heavy global reasoning to every location, we partition a 4K feature map into tiles with content-adaptive selective SSM processing and apply expensive context modeling only to regions that need it.

\subsection{Metadata-based RAW Reconstruction}
Several works study metadata-based raw reconstruction. In this setting, an sRGB image (or preview) is stored with a compact metadata bitstream that is sampled or learned from the raw capture. The raw image is then reconstructed when needed. This setting is related to, but different from InvISP~\cite{xing2021invertible}, which learns an invertible mapping between rendered images and raw signals without explicitly allocating a metadata bitrate. Metadata-based reconstruction instead treats the JPEG/sRGB preview as an available reference and transmits only a compact side bitstream, making rate--distortion efficiency a central objective. Punnappurath and Brown~\cite{punnappurath2021spatially} propose spatially aware metadata sampling and reconstruction. CAM~\cite{nam2022learning} introduces content-adaptive metadata for sRGB-to-raw de-rendering and shows the benefit of online fine-tuning at test time. R2LCM~\cite{wang2023raw} proposes learned compact metadata in feature space and improves entropy modeling for raw image compression; its journal extension Beyond-R2LCM~\cite{wang2024beyond} further enhances the entropy model and achieves stronger rate--distortion performance. These methods provide strong baselines, but their entropy and context models are largely convolution-based, which limits rate--distortion performance due to restricted receptive fields. While Transformer-based approaches can improve performance through long-range modeling, they suffer from low computational efficiency due to quadratic attention complexity and become costly for high-resolution inference.

\subsection{State Space Models for Vision}
State Space Models (SSMs) are an efficient alternative to Transformers for long-sequence modeling. This line of work originates from the Structured State Space sequence (S4) model~\cite{gu2021efficiently}. Mamba~\cite{gu2024mamba} extends SSMs with selective state spaces and input-dependent parameters, which enables linear-time sequence modeling. Recent work also studies theoretical connections between Transformers and SSMs~\cite{dao2024transformers}.
In computer vision, Mamba is adapted into backbones such as Vision Mamba~\cite{zhu2024vision} and VMamba~\cite{liu2024vmamba}. Progress on efficient 2D spatial modeling includes 2DMamba~\cite{zhang20252dmamba}, which proposes a hardware-aware 2D selective scan for gigapixel whole-slide image classification. Other extensions include windowed scanning in LocalMamba~\cite{huang2024localmamba}, hybrid designs such as MambaVision~\cite{hatamizadeh2025mambavision}, and domain-specific variants such as U-Mamba for biomedical segmentation~\cite{ma2024u}. SSM-based models are also applied to image restoration, including MambaIRv2~\cite{guo2025mambairv2} with attentive state-space equations for non-causal modeling, Q-MambaIR~\cite{chen2025q}, and VMambaIR~\cite{shi2025vmambair}.

In image compression and RAW processing, recent works~\cite{zeng2025mambaic, qin2025cassic, qin2026mambasic} explore SSMs for efficient entropy or spatial modeling. RAWMamba~\cite{chen2024rawmamba} uses a Mamba framework for a unified sRGB-to-RAW de-rendering pipeline, while CMIC~\cite{chen2025cmic} proposes a content-adaptive Mamba architecture for learned image compression. These studies demonstrate the promise of SSMs, but they mainly redesign the reconstruction/backbone network or rely on complex token organizations. In contrast, our goal is JPEG-guided metadata-based RAW reconstruction under an explicit metadata bitrate. MambaRaw therefore inserts SSMs into the entropy-parameter network and couples selective computation with a lightweight spatial energy map, so that high-resolution context modeling is concentrated on informative regions while preserving the existing metadata-reconstruction formulation.

\section{Method}

\subsection{Problem Setup}
\label{sec:problem_setup}
\noindent \textbf{Metadata-based Raw Reconstruction.} We study JPEG-guided metadata-based raw reconstruction. Let $\mathbf{x}_{\mathrm{raw}} \in [0,1]^{3\times H\times W}$ denote a scene-referred raw image (in a fixed raw color space) and $\mathbf{x}_{\mathrm{jpg}} \in [0,1]^{3\times H\times W}$ its aligned in-camera JPEG preview. The encoder produces a compact metadata bitstream $\mathbf{s}$ from $(\mathbf{x}_{\mathrm{raw}}, \mathbf{x}_{\mathrm{jpg}})$, and the decoder reconstructs $\hat{\mathbf{x}}$ given $(\mathbf{s}, \mathbf{x}_{\mathrm{jpg}})$. More details can be found in Appendix~\ref{app:detail-network}.

\noindent \textbf{State Space Models.} Our method utilizes the Visual State Space (VSS) block from VMamba~\cite{liu2024vmamba}. In discrete time, the state space model is defined as:
\begin{equation}
h_{t+1} = \mathbf{A}h_t + \mathbf{B}x_t,\quad y_t = \mathbf{C}h_t,
\label{eq:ssm}
\end{equation}
where $x_t, y_t$ are the input and output at time $t$, $h_t$ is the hidden state, and $\mathbf{A,B,C}$ are learned input-dependent parameters. To apply 1D SSMs to 2D feature maps, we employ cross-scan with four scanning directions (left--right, right--left, top--bottom, bottom--top) and merge results:
\begin{equation}
\mathbf{y} = \mathrm{CrossMerge}\big(\mathrm{SS2D}(\mathrm{CrossScan}(\mathbf{x}))\big),
\label{eq:ss2d}
\end{equation}
where $\mathbf{x}, \mathbf{y}$ are the input and output 2D feature maps.

\subsection{Motivation and Overview}
\label{sec:motivation}
Efficiently transmitting and recovering 4K raw images presents a unique challenge: balancing high-fidelity reconstruction with computational feasibility. We address this trade-off between efficiency and effectiveness by identifying two key lightweight modules, both of which can be unified through the lens of spatial feature \textbf{Energy}, defined as the squared magnitude of feature activations $E = f^2$.

\begin{figure}[t]
\centering
\includegraphics[width=\linewidth]{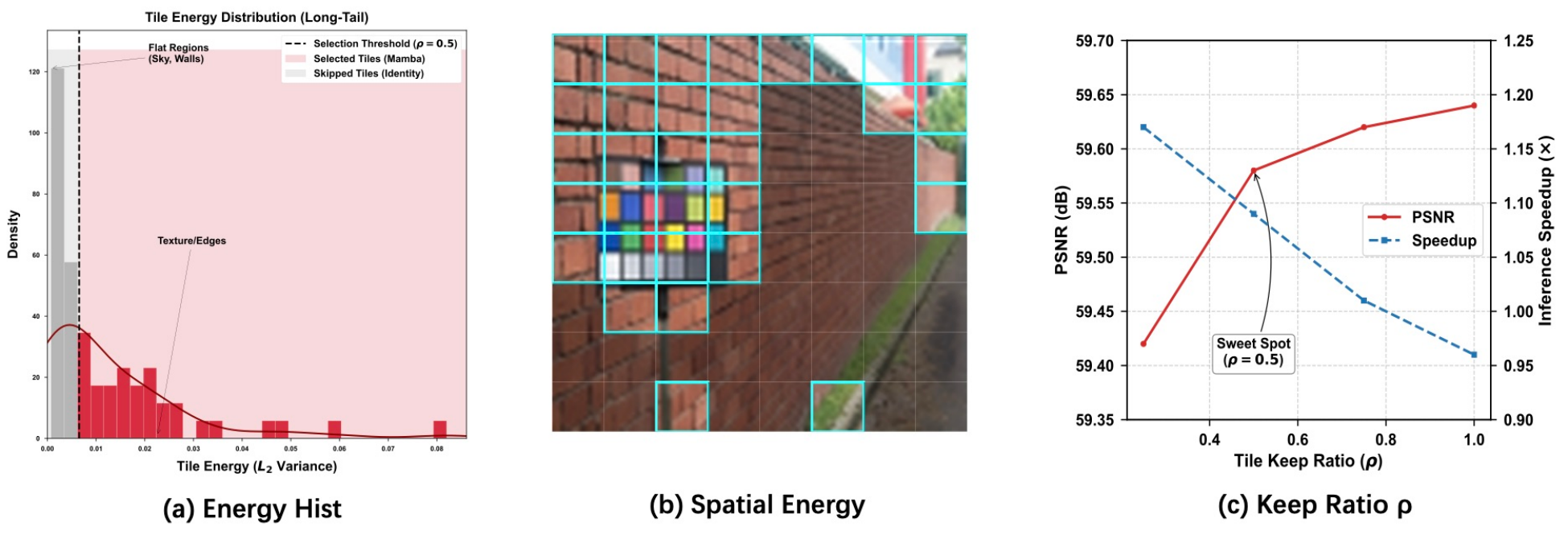}
{
\setlength{\abovecaptionskip}{2pt}
\begin{subfigure}[b]{0.32\linewidth}
\end{subfigure}\hfill
\begin{subfigure}[b]{0.32\linewidth}
\end{subfigure}\hfill
\begin{subfigure}[b]{0.32\linewidth}
\end{subfigure}
}
\caption{Energy analysis and tile selection. \textbf{(a)}: Long-tail energy distribution motivating EAR. \textbf{(b)}: Spatial L2 energy map showing selected high-energy tiles (cyan) at $\rho=0.5$. \textbf{(c)}: Impact of keep ratio $\rho$; $\rho=0.5$ offers the optimal accuracy-speed trade-off.}
\label{fig:tile_analysis}
\end{figure}

\noindent \textbf{Efficiency through Energy Selection.} 
In high-resolution raw images, information is not uniformly distributed. High-energy regions typically correspond to detailed foregrounds (textures, edges), while low-energy regions correspond to smooth backgrounds, as shown in Figure~\ref{fig:tile_analysis}\textcolor{red}{b}. Processing the entire 4K feature map with complex context models is redundant. Therefore, by using energy to distinguish foreground from background, we can selectively apply computationally intensive modeling only where it is most needed.

\noindent \textbf{Effectiveness through Energy Distribution.}
Even within informative regions, the energy distribution is often long-tailed and uneven across spatial locations, as is illustrated in Figure~\ref{fig:tile_analysis}\textcolor{red}{a}. A uniform context model may fail to capture these subtle variations. To maximize effectiveness, we need a mechanism that can refine features adaptively based on their specific energy characteristics, enhancing the representation of complex signals.

Guided by these observations, we propose \textbf{MambaRaw}, which integrates two core modules: (1) \textbf{TileMambaBlock} for efficiency, using tile-wise energy to select and process only information-dense foreground regions; and (2) \textbf{Energy-Aware Refinement (EAR)} for effectiveness, utilizing an identity-initialized residual to align with the intrinsic long-tail energy distribution of raw signals. The overall mixed-scale inference used in our context model is summarized in Algorithm~\ref{alg:tile_mamba}, including both the selective TileMambaBlock processing for efficiency and the EAR refinement (Sec.~\ref{sec:ear}) for energy-calibrated features. 

\subsection{JPEG-Conditioned Reconstruction Backbone}
\begin{figure}[t]
\centering
\includegraphics[width=1\linewidth]{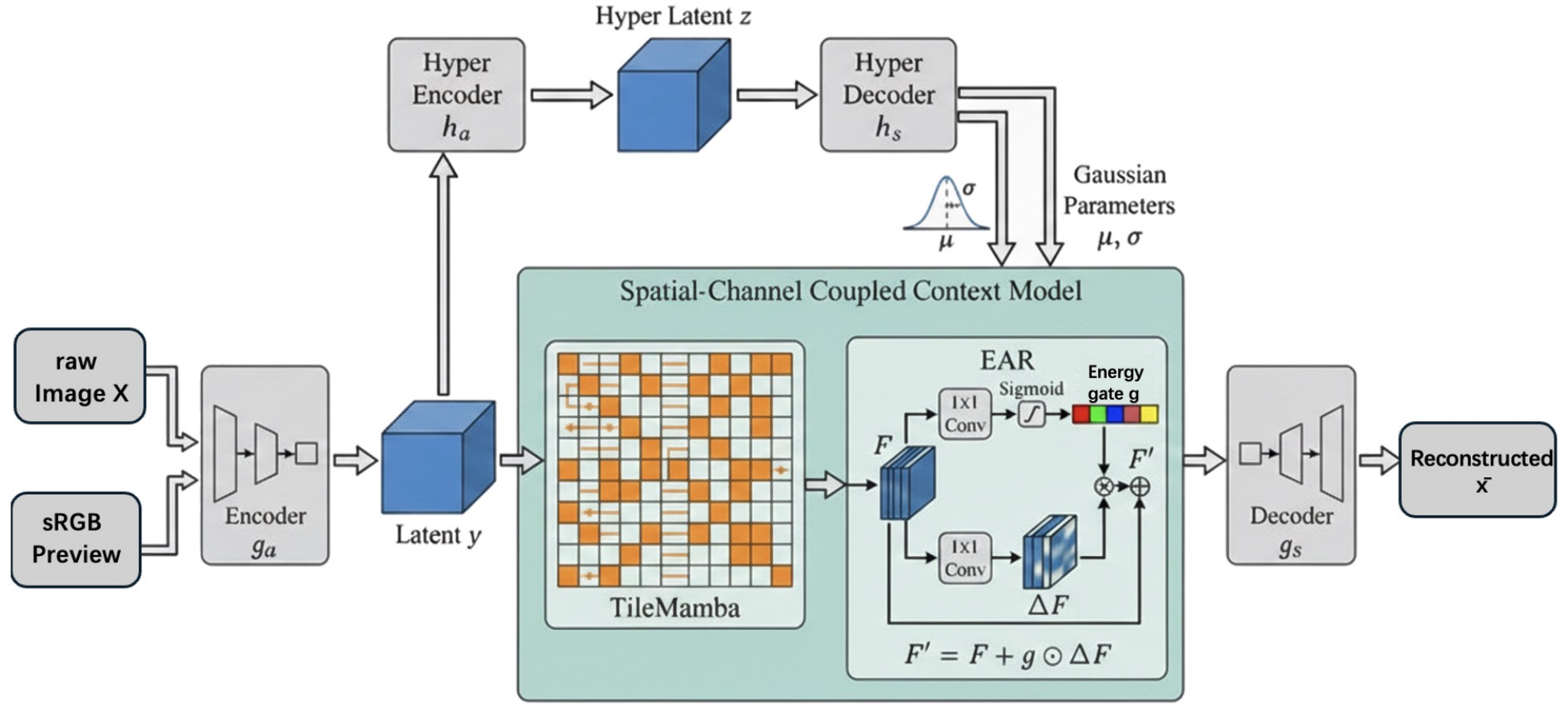}
\caption{The Overall Framework of MambaRaw. We adopt a two-level VAE architecture conditioned on the available JPEG preview. The core innovation lies in the Level-1 Context Model, where we replace standard separate spatial/channel contexts with a coupled design: \textbf{TileMambaBlock} for efficient long-range spatial modeling on selected information-dense tiles, and \textbf{EAR} for lightweight energy-guided refinement.}
\label{fig:framework}
\vskip -5pt
\end{figure}

As shown in Fig.~\ref{fig:framework}, we build on a hyperprior-based framework with learned context modeling following the R2LCM line of work~\cite{wang2023raw,wang2024beyond}. The architecture consists of: (1) a JPEG-conditioned analysis transform $g_a$ that maps $(\mathbf{x}_{\mathrm{raw}}, \mathbf{x}_{\mathrm{jpg}})$ to latents $\mathbf{y}$; (2) hyper analysis/synthesis transforms $h_a, h_s$ that produce hyper-latents $\mathbf{z}$ and side information for entropy modeling; and (3) a JPEG-conditioned synthesis transform $g_s$ that reconstructs $\hat{\mathbf{x}}$ from quantized latents.

We incorporate JPEG guidance via feature concatenation: at each resolution level $l$, we first resize the JPEG preview to match the spatial resolution of layer $l$:
\begin{equation}
\mathbf{x}_{\mathrm{jpg}}^{(l)} = \phi_{l}(\mathbf{x}^{(l-1)}_{\mathrm{jpg}}),
\label{eq:jpeg_resize}
\end{equation}
where $\phi_{l}(\cdot)$ denotes bilinear interpolation to the spatial resolution of layer $l$. We then concatenate the intermediate feature and the resized JPEG feature along channel dimension:
\begin{equation}
\tilde{\mathbf{F}}^{(l)} = \left[\mathcal{M}^{(l-1)}(\tilde{\mathbf{F}}^{(l-1)}), \mathbf{x}_{\mathrm{jpg}}^{(l)}\right],
\label{eq:jpeg_concat}
\end{equation}
where $\mathcal{M}^{(l-1)}$ is the layer $l-1$ mapping function, $[\cdot,\cdot]$ denotes concatenation, and $\tilde{\mathbf{F}}^{(l)}$ is the JPEG-conditioned feature. The same conditioning applies to $g_a, g_s$, and the entropy-parameter branch, injecting JPEG guidance at every scale without cross-attention overhead. Resized JPEG features are concatenated before each mapping, providing running latents and aligned preview cues.

For the entropy model, the conditioned feature first enters an entropy stem ($\psi_{\mathrm{ep}}^{\mathrm{in}}$ mapping to the internal channel dimension $C$) and then passes through the proposed coupled context modules:
\begin{equation}
\mathbf{F}_{\mathrm{in}} = \psi_{\mathrm{ep}}^{\mathrm{in}}(\tilde{\mathbf{F}}),\quad
\mathbf{F}_{c} = \mathrm{TileMambaBlock}(\mathbf{F}_{\mathrm{in}}; T,\rho),\quad
\mathbf{F}' = \mathrm{EAR}(\mathbf{F}_{c}),
\label{eq:context_chain}
\end{equation}
and hyperprior side information:
\begin{equation}
\mathbf{U} = h_s(\hat{\mathbf{z}}),
\label{eq:hyper_side}
\end{equation}
followed by the entropy head $\psi_{\mathrm{ep}}^{\mathrm{out}}$ that predicts independent single Gaussian distribution parameters for each latent element:
\begin{equation}
(\boldsymbol{\mu}, \log \boldsymbol{\sigma}) = \psi_{\mathrm{ep}}^{\mathrm{out}}(\mathbf{F}', \mathbf{U}).
\label{eq:entropy_head}
\end{equation}
Here, $\tilde{\mathbf{F}}$ is the JPEG-conditioned feature, $\mathbf{F}_{\mathrm{in}}$ is the context input before SSM processing, $\mathbf{F}_{c}$ is the TileMambaBlock output, and $\mathbf{F}'$ is the final EAR output.

\subsection{Efficiency: TileMambaBlock with Energy-Guided Selection}
\label{sec:tile_mamba}
To address the efficiency bottleneck of 4K processing, we propose \textbf{TileMambaBlock}, which leverages the spatial distribution of energy to selectively apply long-range context modeling. As discussed in Sec.~\ref{sec:motivation}, high-energy regions ($E=f^2$) effectively distinguish detailed foregrounds from smooth backgrounds. Treating all regions equally with heavy context models incurs unnecessary computation. Instead, TileMambaBlock dynamically identifies and processes only information-dense tiles. This design is fundamentally aligned with the inherent sparsity of high-frequency information in natural images; by restricting advanced scanning mechanisms to structurally complex regions, we drastically reduce required float-point operations (FLOPs) without sacrificing the perceptual fidelity of raw imagery.

\noindent{Tile partition.}
Given the context input $\mathbf{F}_{\mathrm{in}}\in\mathbb{R}^{C\times H\times W}$ in Eq.~\ref{eq:context_chain}, we partition $\mathbf{F}_{\mathrm{in}}$ into non-overlapping tiles of size $T\times T$. The resulting set of tiles is denoted as $\mathcal{T} = \{\mathbf{t}_1, \ldots, \mathbf{t}_{N_t}\}$, where $N_t = \lceil H/T \rceil \times \lceil W/T \rceil$.

To facilitate Energy-Guided Selection, we quantify the information density of each tile using its L2 energy, serving as a computationally inexpensive proxy for entropy:
\begin{equation}
S_i = \frac{1}{C T^2} \sum_{c,h,w} \mathbf{t}_i[c,h,w]^2.
\label{eq:tile_score}
\end{equation}
Based on these scores $S_i$, we identify a subset $\mathcal{S}$ containing the top-$k$ most informative tiles, where $k = \lfloor \rho N_t \rfloor$ is controlled by a keep ratio $\rho \in (0, 1]$. The Mamba-based context modeling is then applied exclusively to these selected tiles:
\begin{equation}
\mathbf{t}'_i =
\begin{cases}
\mathrm{MambaBlock}(\mathbf{t}_i), & i \in \mathcal{S},\\
\mathbf{t}_i, & \text{otherwise},
\end{cases}
\label{eq:tile_selective}
\end{equation}
where $\mathbf{t}'_i$ is the processed tile. This selection reserves expensive context modeling for complex regions. For small patches ($H, W \le T$), it defaults to dense processing.

\subsection{Effectiveness: Energy-Aware Refinement  (EAR)}
\label{sec:ear}
While TileMambaBlock ensures efficiency by selecting important spatial regions, we also need to ensure the effectiveness of context modeling within these regions. Raw image features exhibit a long-tail energy distribution (Figure~\ref{fig:tile_analysis}\textcolor{red}{a}), where information is unevenly distributed across spatial locations and feature dimensions. A static or uniform context model may struggle to adapt to this variance.

To address this, we introduce the \textbf{Energy-Aware Refinement (EAR)}, a lightweight refurbishment module designed to enhance feature representation based on local energy statistics. EAR acts as a spatial-energy refinement step that dynamically adjusts feature responses according to their energy profile. Unlike SENets~\cite{hu2018squeeze} which rely on global average pooling for channel recalibration, EAR explicitly preserves local spatial granularity rather than discarding it via uniform pooling. This introduces a spatially-varying inductive bias that prevents the over-smoothing of critical high-frequency details, allowing the entropy model to adapt robustly to nuanced local raw signal variations. 

\begin{algorithm}[t]
\caption{Mixed-Scale Inference (\textcolor{blue!75}{blue}: TileMambaBlock, \textcolor{red!80}{red}: EAR)}
\label{alg:tile_mamba}
\begin{algorithmic}[1]
\Require JPEG-conditioned feature $\tilde{\mathbf{F}} \in \mathbb{R}^{C_0 \times H \times W}$, tile size $T$, keep ratio $\rho$
\Ensure Refined context features $\mathbf{F}'$
\State \TileKey{$\mathbf{F}_{\mathrm{in}} \gets \psi_{\mathrm{ep}}^{\mathrm{in}}(\tilde{\mathbf{F}})$} \Comment{Context input projection}
\State \TileKey{$N_h \gets \lceil H/T \rceil,\; N_w \gets \lceil W/T \rceil,\; N_t \gets N_hN_w$}
\If{$(H \le T \land W \le T)$ \textbf{or} $(\rho \ge 1)$}
    \State \TileKey{$\mathbf{F}_{c} \gets \text{MambaBlock}(\mathbf{F}_{\mathrm{in}})$} \Comment{Dense fallback}
\Else
    \State \TileKey{Pad $\mathbf{F}_{\mathrm{in}}$ to multiples of $T$: $\mathbf{F}_{\mathrm{in,pad}}$}
    \State \TileKey{Reshape $\mathbf{F}_{\mathrm{in,pad}} \to \{\mathbf{t}_i\}_{i=1}^{N_t}$, $\mathbf{t}_i\in\mathbb{R}^{C\times T\times T}$}
    \State \TileKey{Compute tile score: $S_i \gets \frac{1}{CT^2}\sum \mathbf{t}_i^2$}
    \State \TileKey{$k \gets \max(1,\lfloor \rho N_t \rfloor),\; \mathcal{I}_{\text{top}} \gets \text{TopKIndices}(S,k)$}
    \For{$i \gets 1$ to $N_t$}
        \If{$i \in \mathcal{I}_{\text{top}}$}
            \State \TileKey{$\mathbf{t}'_i \gets \text{MambaBlock}(\mathbf{t}_i)$}
        \Else
            \State \TileKey{$\mathbf{t}'_i \gets \mathbf{t}_i$} \Comment{Skip smooth tiles}
        \EndIf
    \EndFor
    \State \TileKey{Reshape $\{\mathbf{t}'_i\}$ and crop padding: $\mathbf{F}_{c}$}
\EndIf
\State \EARKey{$\mathbf{e} \gets \frac{1}{C}\sum_{j=1}^{C}\mathbf{F}_{c,j}^2$}
\State \EARKey{$\mathbf{g} \gets \sigma(\mathrm{Conv}_{1\times1}(\mathbf{e}))$}
\State \EARKey{$\Delta\mathbf{F} \gets \mathrm{Conv}_{1\times1}\!\left(\mathrm{ReLU}\!\left(\mathrm{Conv}_{1\times1}(\mathbf{F}_{c})\right)\right)$}
\State \EARKey{$\mathbf{F}' \gets \mathbf{F}_{c} + \mathbf{g}\odot\Delta\mathbf{F}$} \Comment{EAR residual}
\State \Return $\mathbf{F}'$
\end{algorithmic}
\end{algorithm}

Equation~\ref{eq:context_chain} and Algorithm~\ref{alg:tile_mamba} outline the path $\tilde{\mathbf{F}}\rightarrow\mathbf{F}_{\mathrm{in}}\rightarrow\mathbf{F}_{c}\rightarrow\mathbf{F}'$. From the context-enhanced feature $\mathbf{F}_{c}\in\mathbb{R}^{C\times H\times W}$, we compute spatial energy $\mathbf{e}$ along channels as an entropy proxy:
\begin{equation}
\mathbf{e} = \frac{1}{C}\sum_{j=1}^C (\mathbf{F}_{c,j})^2 \in \mathbb{R}^{1\times H\times W}.
\label{eq:ear_energy}
\end{equation}
where $\mathbf{F}_{c,j}$ denotes the $j$-th channel slice of $\mathbf{F}_{c}$. We use $\mathbf{e}$ to gate an energy-guided residual for spatial enhancement:
\begin{align}
\mathbf{g} &= \sigma(\mathrm{Conv}_{1\times1}(\mathbf{e})) \in \mathbb{R}^{C\times H\times W}, \label{eq:ear_gate}\\
\Delta \mathbf{F} &= \mathrm{Conv}_{1\times1}\big(\mathrm{ReLU}(\mathrm{Conv}_{1\times1}(\mathbf{F}_{c}))\big), \label{eq:ear_delta}\\
\mathbf{F}' &= \mathbf{F}_{c} + \mathbf{g} \odot \Delta \mathbf{F}, \label{eq:ear_out}
\end{align}
where $\sigma$ is sigmoid, $\odot$ is element-wise multiplication, $\mathbf{g}$ is the gating tensor, and $\Delta\mathbf{F}$ is the residual variation. For stable early training, the last $1\times1$ convolution is zero-initialized, letting EAR start as an identity mapping.

\subsection{Training Strategy}
\noindent{\textbf{Loss function.}}
We optimize the rate--distortion trade-off:
\begin{equation}
\mathcal{L} = R(\hat{\mathbf{y}}) + R(\hat{\mathbf{z}}) + \lambda \cdot D(\mathbf{x}, \hat{\mathbf{x}}),
\label{eq:rd_loss}
\end{equation}
where $R(\cdot)$ denotes estimated bitrate (metadata only) and $D(\cdot)$ is a distortion term. We train models across multiple $\lambda$ values to cover different rate--distortion points.

\noindent{\textbf{Training protocol.}}
We use the same backbone architecture as~\cite{wang2024beyond} and train from scratch. TileMambaBlock and EAR are inserted into the entropy-parameter network. EAR is identity-initialized to ensure stable early-stage training when the entropy model is not yet converged. Tile-wise selection is primarily beneficial for high-resolution inference; for patch-based training where feature maps are small, the module naturally reduces to dense processing.

\section{Experiments}
\subsection{Experimental Setup}
\label{sec:exp_setup}
\noindent \textbf{Datasets.} We evaluate our method on two standard benchmarks: NUS dataset~\cite{cheng2014illuminant} and AdobeFiveK dataset~\cite{bychkovsky2011learning}.
(1) \textbf{NUS dataset}: As a primary benchmark for raw reconstruction, this dataset covers diverse scenes captured by varying sensors. Following the protocol of CAM~\cite{nam2022learning}, we employ three representative subsets (Samsung NX2000, Olympus E-PL6, and Sony SLT-A57) and evaluate on the $4\times$ downsampled version to ensure fair comparison with prior art. (2) \textbf{AdobeFiveK dataset:} We utilize AdobeFiveK to assess reconstruction fidelity under complex lighting and professional photographic conditions. We adopt the Software ISP evaluation setting defined in~\cite{wang2024beyond} with 4,500 training / 500 testing pairs, where sRGB targets are rendered via a software ISP at original resolution.

\noindent \textbf{Implementation details.}
Measurements are performed in the raw-linear color space with inputs normalized to $[0,1]$. We set the tile keep ratio $\rho=0.5$ by default, which balances performance and efficiency. The TileMambaBlock and EAR modules are integrated into the entropy-parameter estimation network. We train the model from scratch using $256\times256$ patches, the Adam optimizer, and mixed precision. Distinct models are trained for each $\lambda \in \{0.02, \allowbreak 0.24, \allowbreak 0.8, \allowbreak 1.5, \allowbreak 2.0, \allowbreak 5.0, \allowbreak 10.0, \allowbreak 20.0\}$ for 1000 epochs.

\noindent \textbf{Metrics and baselines.}
We report PSNR (dB), SSIM, and metadata bitrate (bpp). We compare against baselines including SAM~\cite{punnappurath2021spatially}, CAM~\cite{nam2022learning}, R2LCM~\cite{wang2023raw}, and Beyond-R2LCM~\cite{wang2024beyond}. Detailed definitions of the evaluation metrics are provided in Appendix~\ref{app:details}.

\subsection{Main Results}
\begin{figure}[h!]
\centering
\includegraphics[width=0.9\linewidth]{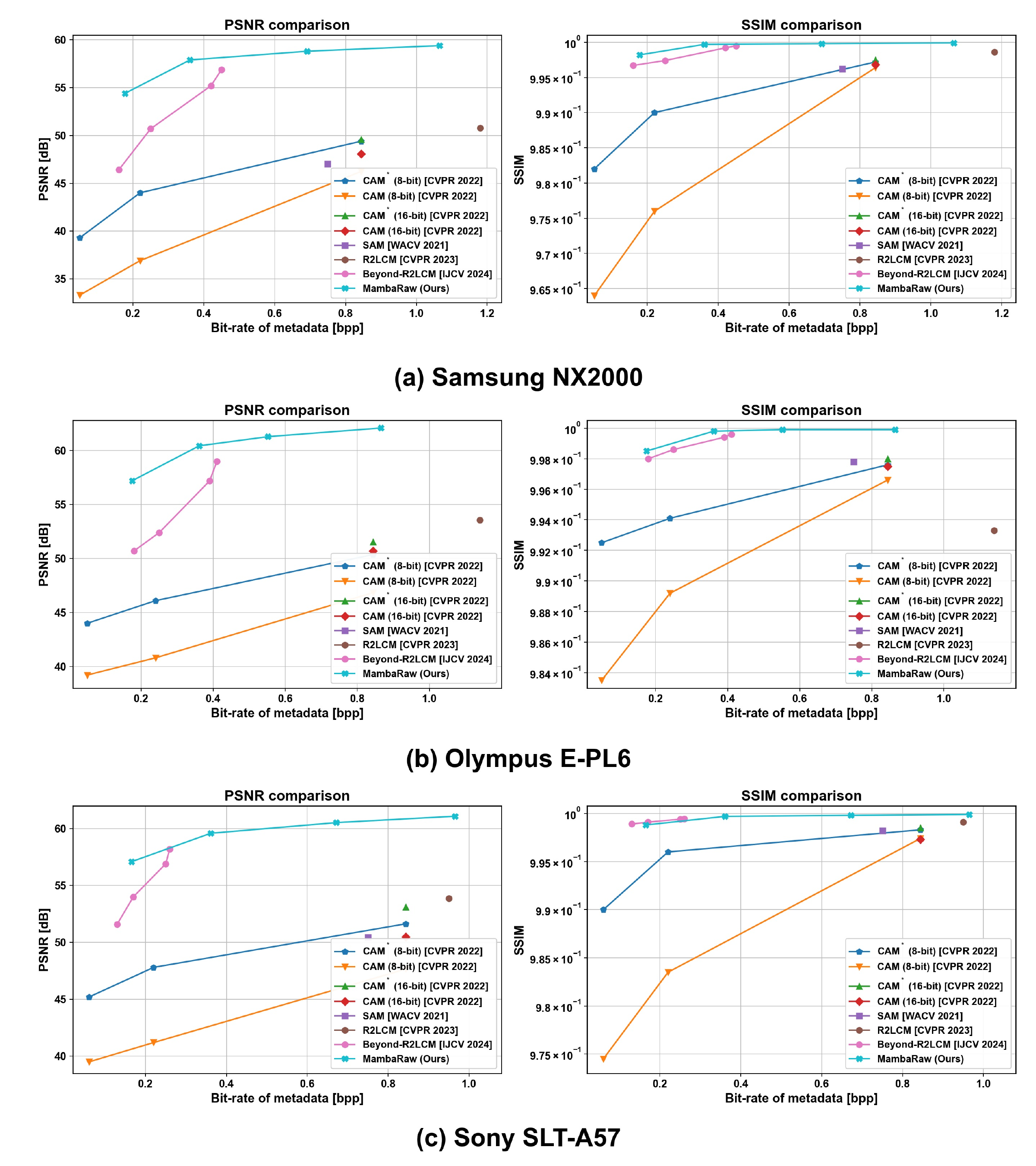}
\caption{RD curves over the NUS dataset (Samsung NX2000, Olympus E-PL6, Sony SLT-A57) following the setting of~\cite{nam2022learning}. The left and right columns report PSNR and SSIM, respectively. For variable-rate models, a single model is trained for each curve and different operating points are obtained by changing the rate--distortion hyper-parameter of the trained model.}
\label{fig:rd_nus}
\end{figure}

\begin{table}[t!]
\centering
\caption{Quantitative results on the NUS dataset. PSNR (dB) and SSIM are evaluated on reconstructed raw images. Reported bpp are metadata-only bitrates, excluding the baseline JPEG preview. The $^*$ marks CAM with test-time online fine-tuning.}
\label{tab:nus}
\scriptsize
\setlength{\tabcolsep}{3.2pt}
\renewcommand{\arraystretch}{1.12}
\resizebox{\linewidth}{!}{\begin{tabular}{l c c c c c c c c}
\toprule
\multirow{2}{*}{\textbf{Method}} & \multirow{2}{*}{\textbf{Input Space}} & \multirow{2}{*}{\textbf{bpp $\downarrow$}} &
\multicolumn{2}{c}{\textbf{Samsung NX2000}} &
\multicolumn{2}{c}{\textbf{Olympus E-PL6}} &
\multicolumn{2}{c}{\textbf{Sony SLT-A57}} \\
\cmidrule(lr){4-5} \cmidrule(lr){6-7} \cmidrule(lr){8-9}
& & &
\textbf{PSNR $\uparrow$} & \textbf{SSIM $\uparrow$} &
\textbf{PSNR $\uparrow$} & \textbf{SSIM $\uparrow$} &
\textbf{PSNR $\uparrow$} & \textbf{SSIM $\uparrow$} \\
\midrule
SAM~\cite{punnappurath2021spatially}                      & 16-bit & 0.7500 & 47.03 & 0.9962 & 49.35 & 0.9978 & 50.44 & 0.9982 \\
CAM~\cite{nam2022learning}        & 8-bit  & 0.8438 & 46.24 & 0.9964 & 46.84 & 0.9966 & 47.66 & 0.9974 \\
CAM~\cite{nam2022learning}       & 16-bit & 0.8438 & 48.08 & 0.9968 & 50.71 & 0.9975 & 50.49 & 0.9973 \\
CAM$^*$~\cite{nam2022learning} & 8-bit  & 0.8438 & 49.40 & 0.9972 & 50.37 & 0.9976 & 51.63 & 0.9983 \\
CAM$^*$~\cite{nam2022learning} & 16-bit & 0.8438 & 49.57 & 0.9975 & 51.54 & 0.9980 & 53.11 & 0.9985 \\
R2LCM~\cite{wang2023raw}                          & 8-bit  & 1.2250 & 50.78 & 0.9986 & 53.56 & 0.9993 & 53.87 & 0.9991 \\
Beyond-R2LCM~\cite{wang2024beyond}                  & 8-bit  & 0.3763 & 56.74 & 0.9996 & 59.04 & 0.9997 & 58.21 & 0.9996 \\
\textbf{MambaRaw (Ours)}                            & 8-bit  & \textbf{0.3612} & \textbf{57.91} & \textbf{0.9997} & \textbf{60.45} & \textbf{0.9998} & \textbf{59.58} & \textbf{0.9997} \\
\bottomrule
\end{tabular}}
\end{table}

\begin{table}[t]
\centering
\caption{Quantitative evaluation on AdobeFiveK dataset. The sRGB input is rendered using a software ISP and spatial resolution remains the same as the original raw image. The reported bits per pixel (bpp) are metadata-only bitrates.}
\label{tab:fivek}
\setlength{\tabcolsep}{8pt}
\resizebox{0.7\linewidth}{!}{\begin{tabular}{c c c c}
\toprule
\textbf{Method} & \textbf{bpp} & \textbf{PSNR} & \textbf{SSIM} \\
\midrule
InvISP~\cite{xing2021invertible} & N/A & 52.69 & 0.9994 \\
SAM~\cite{punnappurath2021spatially} & 9.566e-4 & 49.61 & 0.9987 \\
SAM~\cite{punnappurath2021spatially} & 9.521e-3 & 54.76 & 0.9995 \\
CAM~\cite{nam2022learning}  & 8.438e-1 & 56.72 & 0.9996 \\
R2LCM~\cite{wang2023raw} (w/o metadata) & N/A & 53.03 & 0.9993 \\
\midrule
R2LCM~\cite{wang2023raw} & 4.901e-4 & 58.14 & 0.9997 \\
Beyond-R2LCM~\cite{wang2024beyond} & 3.760e-4 & 58.44 & 0.9997 \\
\textbf{Ours} & \textbf{3.150e-4} & \textbf{58.55} & \textbf{0.9997} \\
\midrule
R2LCM~\cite{wang2023raw} & 1.045e-2 & 59.02 & 0.9994 \\
Beyond-R2LCM~\cite{wang2024beyond} & 2.916e-3 & 59.09 & 0.9997 \\
\textbf{Ours} & \textbf{2.450e-3} & \textbf{59.18} & \textbf{0.9998} \\
\bottomrule
\end{tabular}}
\vskip -0.2in
\end{table}

\noindent \textbf{Rate--distortion trade-off.} Figure~\ref{fig:rd_nus} and Table~\ref{tab:nus} present the rate--distortion comparison on the NUS dataset. Across all three camera subsets (Samsung, Olympus, Sony), MambaRaw consistently dominates the rate-distortion frontier. By effective tile-wise selective scanning and energy-guided refinement, our method delivers superior reconstruction quality (PSNR and SSIM) at comparable or lower metadata bitrates. In particular, our method outperforms robust metadata-based baselines such as R2LCM and Beyond-R2LCM by significant margins. Specifically, on the Samsung subset, MambaRaw achieves a \textbf{1.2 dB} PSNR gain, while on the Sony and Olympus subsets, the improvement reaches up to \textbf{1.4 dB}, particularly in detailed texture regions where conventional models struggle. This robust superiority across diverse sensor statistics and bitrates confirms that our spatial-energy coupled design effectively resolves the bottleneck of high-resolution context modeling.

\noindent \textbf{Quantitative comparison.} We also extend performance evaluation to the AdobeFiveK dataset, which introduces greater variability in scene content and lighting. Table~\ref{tab:fivek} summarizes the quantitative comparison against state-of-the-art methods. Notably, in the challenging low-bitrate regime ($<$ 5e-4 bpp), MambaRaw demonstrates superior efficiency, surpassing the PSNR of Beyond-R2LCM by \textbf{0.11} dB (58.55 vs. 58.44 dB) while requiring approximately \textbf{16\% fewer bits} (3.150e-4 vs. 3.760e-4 bpp). This strict rate--distortion advantage is critical for applications where bandwidth is constrained.

\noindent \textbf{Performance at Target Resolution (4K) and Efficiency Analysis.}
In addition to rate distortion results on $4\times$ downsampled benchmarks, MambaRaw is designed for practical high resolution (4K) settings. To test whether the accuracy gains persist at the target resolution, we report full resolution performance ($3840 \times 2160$) in Table~\ref{tab:4k_efficiency}. In true 4K, our method improves PSNR by \textbf{1.37\,dB} over the baseline without increasing the parameter count. By performing selective scanning only on information dense tiles, our Spatial Energy Coupled Context Modeling avoids redundant global computation. As a result, MambaRaw reduces FLOPs by about \textbf{56\%} and lowers end to end wall clock latency by 9\% compared with Beyond R2LCM. At 4K, the CNN baseline exhibits substantial memory growth and approaches the device limit ($22.8$\,GB), whereas MambaRaw remains within a consumer level memory budget ($10.2$\,GB).

\subsection{Ablation Study}
We perform comprehensive ablation studies to validate our design choices. All experiments are conducted on Sony SLT-A57 at $\lambda=0.8$ unless otherwise stated.

\begin{table}[t]
\centering
\small
\begin{minipage}[t]{0.49\linewidth}
\centering
\begin{minipage}[t][5.5em][t]{\linewidth}
\captionsetup{type=table}
\caption{Performance and efficiency on \emph{true} 4K resolution inputs ($3840{\times}2160$, $\lambda{=}0.8$). MambaRaw retains superior accuracy with high efficiency.}
\label{tab:4k_efficiency}
\end{minipage}
\end{minipage}%
\hfill%
\begin{minipage}[t]{0.49\linewidth}
\centering
\begin{minipage}[t][5.5em][t]{\linewidth}
\captionsetup{type=table}
\caption{Progressive analysis of component effectiveness. This study validates the individual structural contribution of each proposed module.}
\label{tab:components}
\end{minipage}
\end{minipage}

\begin{minipage}[t]{0.49\linewidth}
\centering
\setlength{\tabcolsep}{1.5pt}
\renewcommand{\arraystretch}{1.45}
\resizebox{\linewidth}{!}{\begin{tabular}[t]{l c c c c c}
\toprule
\textbf{Method} & \textbf{FLOPs (G)}$\downarrow$ & \textbf{Mem (GB)}$\downarrow$ & \textbf{Time (ms)}$\downarrow$ & \textbf{PSNR}$\uparrow$ & \textbf{SSIM}$\uparrow$ \\
\midrule
Baseline~\cite{wang2024beyond} & 5420.7 & 22.8 & 3125 & 55.21 & 0.9982 \\
\textbf{Ours} & \textbf{2380.5} & \textbf{10.2} & \textbf{2859} & \textbf{56.58} & \textbf{0.9991} \\
\bottomrule
\end{tabular}}
\end{minipage}%
\hfill%
\begin{minipage}[t]{0.49\linewidth}
\centering
\setlength{\tabcolsep}{3.5pt}
\renewcommand{\arraystretch}{0.87}
\resizebox{\linewidth}{!}{\begin{tabular}[t]{l c c c c c c}
\toprule
\textbf{Method} & \textbf{EAR} & \textbf{SSM} & \textbf{TileSelect} & \textbf{PSNR} $\uparrow$ & \textbf{SSIM} $\uparrow$ & \textbf{Time} $\downarrow$ \\
\midrule
Baseline & - & - & - & 58.21 & 0.9996 & 563 \\
+ EAR & \checkmark & - & - & 58.55 & 0.9996 & 570 \\
+ SSM (Dense) & \checkmark & \checkmark & - & 59.61 & 0.9997 & 584 \\
\textbf{Ours} & \checkmark & \checkmark & \checkmark & \textbf{59.58} & \textbf{0.9997} & \textbf{515} \\
\bottomrule
\end{tabular}}
\end{minipage}
\end{table}

\noindent \textbf{Effectiveness of Individual Components.}
We analyze the contribution of each component through progressive integration.
Results in Table~\ref{tab:components} show the step-by-step improvements. Specifically, the metadata-based baseline reaches 58.21 dB. Adding the Energy-Aware Refinement (EAR) brings a steady gain (0.34 dB) with negligible latency increase, ensuring robust spatial-energy adaptation. Integrating the dense SSM context model yields a massive performance promotion (1.06 dB) due to superior global spatial modeling, but slightly increases latency to 584 ms. Finally, employing the tile-wise selection mechanism (TileMambaBlock) maintains the high performance (59.58 dB) and reduces inference time by 12\% (584 $\to$ 515 ms), proving that selective processing successfully prunes redundancy.

\begin{table}[t]
\centering
\small
\begin{minipage}[t]{0.49\linewidth}
\centering
\begin{minipage}[t][4.8em][t]{\linewidth}
\captionsetup{type=table}
\caption{Comparison of different tile selection metrics. L2 Energy provides the best balance of accuracy and speed.}
\label{tab:ablation_metric}
\end{minipage}%
\setlength{\tabcolsep}{3.5pt}
\renewcommand{\arraystretch}{1.02}
\vtop{\hbox{\resizebox{0.95\linewidth}{!}{\begin{tabular}{lccc}
\toprule
\textbf{Metric} & \textbf{PSNR} $\uparrow$ & \textbf{SSIM} $\uparrow$ & \textbf{Total Time (ms)} $\downarrow$ \\
\midrule
Random & 59.25 & 0.9996 & 515 \\
Entropy & 59.55 & 0.9997 & 542 \\
Gradient & 59.52 & 0.9997 & 528 \\
\textbf{L2 Energy (Ours)} & \textbf{59.58} & \textbf{0.9997} & \textbf{515} \\
\bottomrule
\end{tabular}}}}
\end{minipage}%
\hfill%
\begin{minipage}[t]{0.49\linewidth}
\centering
\begin{minipage}[t][4.8em][t]{\linewidth}
\captionsetup{type=table}
\caption{Comparison of foundational context modeling blocks. SSM offers the best performance-speed trade-off.}
\label{tab:foundational}
\end{minipage}%
\setlength{\tabcolsep}{3.5pt}
\renewcommand{\arraystretch}{1.35}
\vtop{\hbox{\resizebox{\linewidth}{!}{\begin{tabular}{l c c c}
\toprule
\textbf{Block Type} & \textbf{PSNR} $\uparrow$ & \textbf{SSIM} $\uparrow$ & \textbf{Total Time (ms)} $\downarrow$ \\
\midrule
CNN (ResBlock) & 58.45 & 0.9996 & \textbf{480} \\
Transformer (Win-Attn) & 59.52 & 0.9997 & 620 \\
\textbf{SSM (Ours)} & \textbf{59.58} & \textbf{0.9997} & 515 \\
\bottomrule
\end{tabular}}}}
\end{minipage}
\end{table}

\noindent \textbf{Effectiveness of Tile Selection Metric.}
We justify our choice of L2 Energy as the tile selection metric by comparing it with random selection, Entropy, and Gradient Magnitude. As shown in Table~\ref{tab:ablation_metric}, {Random} selection leads to a noticeable performance drop (59.25 dB). While Entropy and Gradient metrics achieve competitive performance, they incur additional computational overhead. By contrast, \textbf{L2 Energy} achieves the highest efficiency (515 ms) with comparable SOTA performance. Figure~\ref{fig:tile_analysis}\textcolor{red}{b} visualizes the spatial energy distribution: high-energy tiles concentrate on edges and textures while ignoring smooth background regions, validating the rationale for selective processing.

\noindent \textbf{Impact of Tile Keep Ratio $\rho$.}
We analyze the trade-off between performance and efficiency by varying the tile keep ratio $\rho$. As shown in Figure~\ref{fig:tile_analysis}\textcolor{red}{c}, increasing $\rho$ scans more regions, slightly improving reconstruction quality with lower inference speedup. However, PSNR gains saturate beyond $\rho=0.5$. We select $\rho=0.5$ as a widely applicable default, achieving a sweet spot that maintains state-of-the-art results with significant speedup.

\noindent \textbf{Impact of Foundational Models.}
To verify the effectiveness of the proposed SSM-based design, we replace the core TileMambaBlock with CNN-based and Transformer-based (Swin Transformer~\cite{liu2021swin}) alternatives.
As shown in Table~\ref{tab:foundational}, the CNN variant suffers from limited long-range modeling with poor reconstruction performance. The Transformer variant matches our performance but incurs 20\% higher latency due to quadratic attention complexity. By contrast, our SSM-based design achieves the best rate-distortion performance with optimal efficiency, strongly showing better global context modeling of the selective scanning in 4K image reconstruction.

\subsection{Qualitative Visualization}
\begin{figure}[t]
\centering
\includegraphics[width=1.0\linewidth]{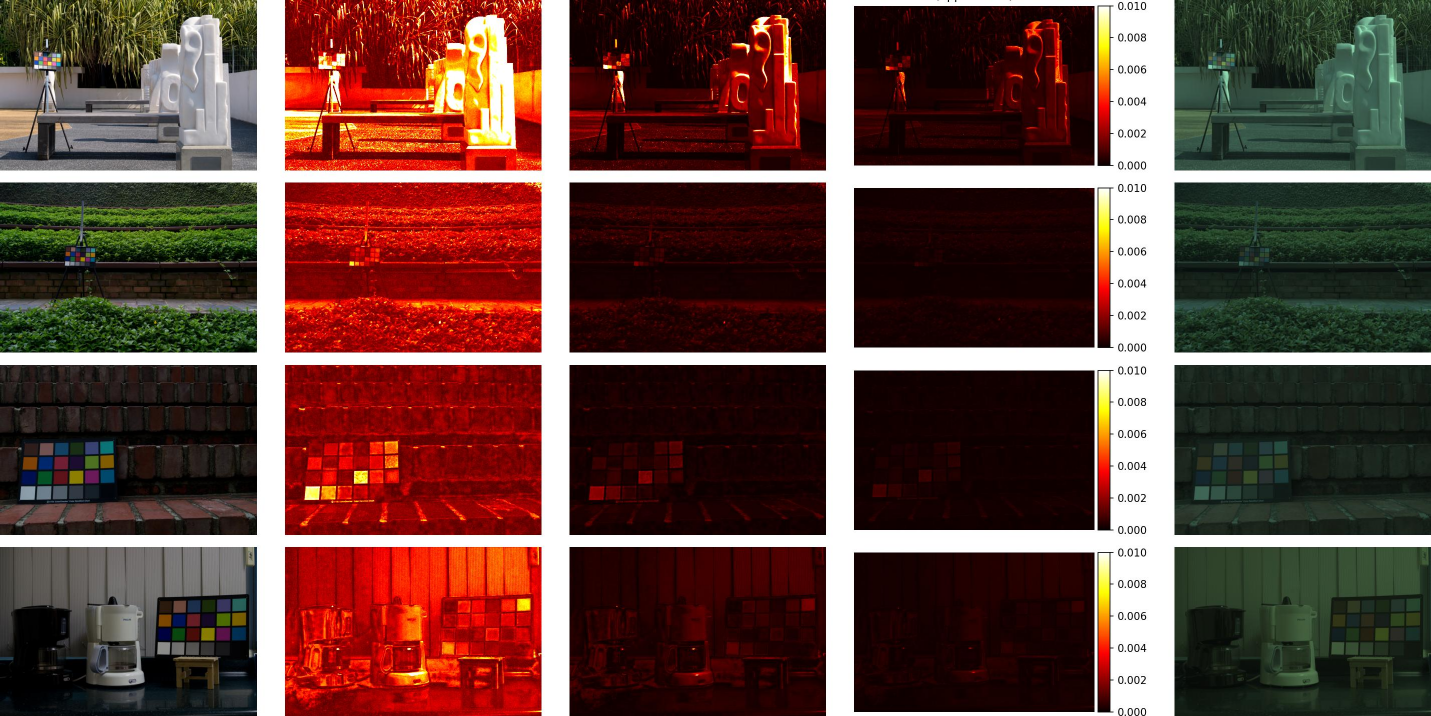}
\QualLegendRow
\caption{Qualitative comparison on Sony SLT-A57. Error maps show the per-pixel maximum absolute error over the three channels (after gamma correction for visibility); darker indicates smaller error.}
\label{fig:vis_sony}
\end{figure}
To understand the superiority of MambaRaw, we present a qualitative comparison of the per-pixel absolute error in Figure~\ref{fig:vis_sony}. The sRGB preview (a) shows scenes with complex high-frequency textures (\eg, the text on the book spine and the fabric pattern). The baseline Beyond-R2LCM (c) struggles to align with these fine details, resulting in residual errors in the error map (brighter regions). In contrast, MambaRaw (d) effectively suppresses these errors by refined spatial energy enhancement. Concretely, coupled spatial-channel context modeling allows the network to better predict complex signal variations, leading to a darker error map that indicates higher reconstruction fidelity. This aligns with our quantitative results, confirming that selective SSM processing captures critical structural information more effectively than existing methods.

\FloatBarrier
\section{Conclusion}
In this paper, we present \textbf{MambaRaw}, a JPEG-conditioned learned framework for efficient metadata-based 4K raw image reconstruction. We 
integrate state space models into entropy parameter estimation, and propose spatial-energy coupled context modeling with two lightweight modules: TileMambaBlock performs tile-wise selective context modeling to enable practical high-resolution inference and EAR uses energy-guided refinement to improve entropy enhancement and enhance feature representation to match the long-tail energy distribution of raw signals while maintaining stable training. Experiments on three camera datasets show consistent rate--distortion gains over strong metadata-based baselines, including up to 1.4\,dB PSNR at similar bitrates, and reduce end-to-end coding latency by 9\% on average.

\noindent \textbf{Future work.}
This work focuses on single-frame raw reconstruction. A natural extension is raw video processing that exploits temporal redundancy. This direction can incorporate temporal SSM designs such as VideoMamba~\cite{li2024videomamba}. Another direction is hardware-aware optimization that maps selective SSM processing to mobile accelerators for real-time computational photography.

%
%
\bibliographystyle{splncs04}
\bibliography{main}

\newpage
\appendix 
\setcounter{table}{0}
\renewcommand{\thetable}{A\arabic{table}}
\setcounter{figure}{0}
\renewcommand{\thefigure}{A\arabic{figure}}
\section{More Details}
\label{app:details}
\subsection{Network Architecture}
\label{app:detail-network}
Our MambaRaw framework directly adopts the two-level JPEG-conditioned
learned-context backbone of Beyond-R2LCM~\cite{wang2024beyond}. It consists
of two cascaded analysis transforms and their corresponding synthesis
transforms. At both levels, latent elements are progressively coded using
learned spatial sampling masks and JPEG-conditioned context prediction.
We preserve the original backbone and replace only the Level-1
entropy-parameter network with an input projection, TileMambaBlock, EAR,
and an output projection. Detailed architectural settings are summarized
in Table~\ref{tab:appendix_hparams}.

\begin{table}[h]
\vskip -0.2 in
\centering
\small
\setlength{\tabcolsep}{6pt}
\renewcommand{\arraystretch}{1.15}
\caption{Key hyperparameters used in our implementation.}
\label{tab:appendix_hparams}
\begin{tabular}{l c}
\toprule
\textbf{Component} & \textbf{Setting} \\
\midrule
Backbone channel width & $N=192$ \\
Latent channel reduction factor & $8$ \\
Reconstruction levels & $2$ \\
Learned spatial sampling rounds & $4$ \\
Context prediction & Masked deconvolution \\
Entropy distribution & Independent Gaussian $(\mu,\sigma)$ \\
Tile size (latent) & $T=64$ \\
Tile keep ratio & $\rho=0.5$ (default) \\
SSM block & VSS (VMamba) with expansion factor 2 \\
EAR initialization & last $1\times1$ conv zero-initialized \\
\bottomrule
\end{tabular}
\vskip -0.2in
\end{table}

\noindent \textbf{Entropy model integration.}
The core innovation lies in the Level-1 entropy-parameter network. We
replace its original convolutional parameter estimator with an input
projection, \textbf{TileMambaBlock}, \textbf{Energy-Aware Refinement
(EAR)}, and an output projection. The same entropy-parameter network is
used during training, encoding, and decoding.

\noindent \textbf{JPEG-conditioned feature injection.}
For the Level-1 entropy-parameter network, we concatenate the feature
propagated from the deeper level, the progressively predicted context
feature, and the cumulative sampling mask. The JPEG preview is
bilinearly resized to the corresponding latent resolution and
concatenated before the input projection. The resulting feature follows
the chain in Eq.~\ref{eq:context_chain}, i.e.,
$\tilde{\mathbf{F}}\rightarrow\mathbf{F}_{\mathrm{in}}
\rightarrow\mathbf{F}_{c}\rightarrow\mathbf{F}'$.
The aligned JPEG feature is additionally concatenated in the output
projection, which predicts the Gaussian mean and scale $(\mu,\sigma)$.

\noindent \textbf{Context Modules.}
\textbf{(1) TileMambaBlock.} Given an intermediate feature map $\mathbf{F}\in\mathbb{R}^{C\times H\times W}$, we partition it into non-overlapping $T\times T$ tiles (on the latent feature resolution). We score each tile by its L2 energy (Eq.~\ref{eq:tile_score}) and apply the SSM-based context block only to the top-$k$ tiles, where $k=\lfloor \rho N_t\rfloor$. Unless otherwise specified, we use $T=64$ and $\rho=0.5$.

\noindent \textbf{(2) State space block.} The internal SSM uses the Visual State Space (VSS) block from VMamba~\cite{liu2024vmamba} with a state expansion factor of 2 and 2D cross-scan (four directions). This provides long-range spatial aggregation with linear-time complexity.The VSS and selective-scan implementation follows
MambaIC~\cite{zeng2025mambaic}, while the entropy-coding structure
follows Beyond-R2LCM~\cite{wang2024beyond}.

\noindent \textbf{(3) EAR.} EAR refines entropy features based on local energy statistics while preserving spatial granularity. Given features $\mathbf{F}$, we compute an energy map (Eq.~\ref{eq:ear_energy}) and predict a gating tensor (Eq.~\ref{eq:ear_gate}) to modulate a lightweight residual branch (Eqs.~\ref{eq:ear_delta}--\ref{eq:ear_out}).

\noindent \textbf{(4) Learned spatial sampling.}
Following Beyond-R2LCM~\cite{wang2024beyond}, latent elements are progressively coded in four spatial sampling rounds. In each round, a JPEG-conditioned learned mask selects the current spatial positions. The context-prediction network uses the previously reconstructed positions, the cumulative sampling mask, and the aligned JPEG preview to estimate the Gaussian mean and scale. The same sampling order is used during training, encoding, and decoding.

\noindent \textbf{What we mean by 4K.}
Throughout the paper, ``4K'' refers to \emph{4K-class} high-resolution RAW captures (\ie, images whose long side is on the order of $\sim$4K pixels). Note that some benchmarks adopt downsampled evaluation protocols for fair comparison: for NUS, we report RD results on the $4\times$ downsampled setting (Sec.~\ref{sec:exp_setup}). Our efficiency design targets the entropy model bottleneck on large feature maps and is therefore most beneficial at higher resolutions; the relative speedup from tile-wise selection typically increases with input size.

\subsection{Training Strategy}
We optimize all models using Adam with hyperparameters $(\beta_1,\beta_2)=(0.9,0.999)$. We set the initial learning rate to $1\times10^{-4}$ for all parameters. We use a cosine annealing schedule to decay the learning rate to $1\times10^{-6}$ over 1000 epochs, which improves stability in the later stage of training. We train all models from scratch with a total batch size of 8 on NVIDIA RTX A30 GPUs. We adopt Automatic Mixed Precision (AMP) to reduce memory usage and improve training throughput while maintaining reconstruction quality. During training, we apply data augmentation on the fly to $256 \times 256$ patches, including random horizontal flips, random vertical flips, and random $90^{\circ}$ rotations. This augmentation reduces overfitting and improves generalization of our spatial energy coupled context modeling across diverse raw image sequences.

\subsection{Efficiency Measurement Details}
Table~\ref{tab:latency_breakdown} breaks down the runtime of TileMambaBlock. The selection overhead from L2 scoring, Top-$K$, and padding/reshaping is only 27 ms (5.2\%), showing that the speedup mainly comes from avoiding dense SSM scanning on low-information tiles.

\begin{table}[t]
\centering
\small
\setlength{\tabcolsep}{4pt}
\renewcommand{\arraystretch}{1.05}
\caption{Tile-selection overhead on Sony SLT-A57. Runtime is measured for the entropy-context branch at $\lambda=0.8$.}
\label{tab:latency_breakdown}
\resizebox{0.6\linewidth}{!}{%
\begin{tabular}{lcc}
\toprule
\textbf{Item} & \textbf{Time (ms)} & \textbf{Share} \\
\midrule
Dense SSM w/o selection & 584 & -- \\
TileMamba w/ selection & 515 & 100\% \\
\midrule
L2 score computation & 10 & 1.9\% \\
Top-$K$ selection & 6 & 1.2\% \\
Padding/reshape & 11 & 2.1\% \\
Selection overhead total & 27 & 5.2\% \\
Selected MambaBlock scan & 305 & 59.2\% \\
Other codec modules & 183 & 35.5\% \\
\bottomrule
\end{tabular}}
\end{table}

\subsection{Evaluation Metrics}
We use two standard full reference image quality metrics to evaluate reconstruction fidelity: Peak Signal to Noise Ratio (PSNR) and the Structural Similarity Index Measure (SSIM). PSNR quantifies the pixel level difference between the reconstructed raw image and the ground truth, and it is reported in decibels (dB). Higher PSNR indicates lower distortion. SSIM measures similarity in structural information, luminance, and contrast, and it ranges from 0 to 1. A value of 1 indicates perfect structural agreement. We compute both metrics on raw linear RGB images normalized to the $[0,1]$ range.

\begin{table}[t]
\centering
\small
\setlength{\tabcolsep}{5pt}
\renewcommand{\arraystretch}{1.1}
\caption{NUS subsets used in this work and their spatial resolution protocol.}
\label{tab:nus_appendix_details}
\begin{tabular}{lcc}
\toprule
\textbf{Camera Subset} & \textbf{Native RAW Resolution} & \textbf{$4\times$ Evaluation Resolution} \\
\midrule
Samsung NX2000 & $5472\times3648$ & $1368\times912$ \\
Olympus E-PL6 & $4608\times3456$ & $1152\times864$ \\
Sony SLT-A57 & $4912\times3264$ & $1228\times816$ \\
\bottomrule
\end{tabular}
\end{table}

\subsection{Dataset Protocol Details (NUS)}
To make the NUS evaluation protocol explicit, we summarize the exact setup used in this paper and in the compared metadata-based baselines (see Table~\ref{tab:nus_appendix_details}):
\begin{itemize}
    \item \textbf{Subset cameras.} Samsung NX2000, Olympus E-PL6, and Sony SLT-A57.
    \item \textbf{Input-target pair.} Aligned in-camera sRGB preview (input) and corresponding RAW image (target).
    \item \textbf{Resolution protocol.} Following Nam \emph{et al.}~\cite{nam2022learning}, we evaluate on the $4\times$ downsampled release to ensure fair comparison with prior work.
    \item \textbf{Split protocol.} We follow the official processed split adopted in Beyond-R2LCM~\cite{wang2024beyond} and do not re-split the data.
    \item \textbf{Color space and normalization.} All measurements are performed in raw-linear space with values normalized to $[0,1]$.
\end{itemize}

\section{More Experimental Results}
Before the camera-wise ablations, Table~\ref{tab:tile_ablation} reports the sensitivity to tile size and keep ratio. Tables~\ref{tab:ablation_sony}--\ref{tab:ablation_olympus} then provide detailed ablations on three camera subsets at representative rate--distortion operating points ($\lambda \in \{0.02, 0.8, 5.0, 20.0\}$).

\subsection{Hyperparameter Sensitivity}
Table~\ref{tab:tile_ablation} analyzes the sensitivity of the two key hyperparameters in TileMambaBlock: the tile size $T$ and the keep ratio $\rho$. For tile size, smaller tiles ($T=16,32$) give slightly higher PSNR but incur more overhead, while $T=128$ is faster but less accurate due to overly coarse energy-based selection; thus, $T=64$ offers a better balance. For keep ratio, dense processing ($\rho=1.0$) only improves PSNR over $\rho=0.5$ by 0.03 dB but costs much more latency, whereas $\rho=0.25$ is faster but loses informative tiles, so we use $T=64$ and $\rho=0.5$ as the default near-dense-accuracy and low-latency configuration.

\begin{table}[t]
\centering
\small
\setlength{\tabcolsep}{4pt}
\renewcommand{\arraystretch}{1.05}
\caption{Tile-size and Top-$K$/keep-ratio sensitivity on Sony SLT-A57.}
\label{tab:tile_ablation}
\resizebox{0.6\linewidth}{!}{%
\begin{tabular}{llccc}
\toprule
\textbf{Study} & \textbf{Setting} & \textbf{PSNR} & \textbf{SSIM} & \textbf{Time (ms)} \\
\midrule
\multirow{4}{*}{Tile size} & $T=16$ & 59.60 & 0.9997 & 544 \\
 & $T=32$ & 59.59 & 0.9997 & 528 \\
 & $T=64$ & 59.58 & 0.9997 & 515 \\
 & $T=128$ & 59.50 & 0.9996 & 507 \\
\midrule
\multirow{4}{*}{Keep ratio} & $\rho=0.25$ & 59.42 & 0.9996 & 482 \\
 & $\rho=0.50$ & 59.58 & 0.9997 & 515 \\
 & $\rho=0.75$ & 59.60 & 0.9997 & 556 \\
 & $\rho=1.00$ & 59.61 & 0.9997 & 584 \\
\bottomrule
\end{tabular}}
\end{table}

\subsection{Detailed Ablations Across Camera Subsets}
Across all three camera subsets, we observe consistent trends: (i) dense SSM + EAR yields strong RD gains but increases latency; (ii) enabling tile-wise selection retains most of the RD improvements while substantially reducing total coding time; and (iii) the relative improvements remain stable from low-rate to high-rate regimes, indicating that the proposed modules are not tuned to a single operating point.

\begin{table}[!htbp]
\scriptsize
\centering
\setlength{\tabcolsep}{2.8pt}
\renewcommand{\arraystretch}{1.05}
\caption{Detailed ablation on Sony SLT-A57 across all $\lambda$ values.}
\label{tab:ablation_sony}
\begin{tabular}{cccccccc}
\toprule
\textbf{$\lambda$} & \textbf{Config} & \textbf{PSNR (dB)} & \textbf{SSIM} & \textbf{bpp} & \textbf{Enc. (ms)} & \textbf{Dec. (ms)} & \textbf{Total (ms)} \\ 
\midrule
0.02 & Baseline & 56.45 & 0.9986 & 0.172 & 167 & 321 & 488 \\
0.02 & + SSM \& EAR & 57.12 & 0.9989 & 0.168 & 182 & 340 & 522 \\
0.02 & MambaRaw & 57.08 & 0.9988 & 0.165 & 155 & 302 & 457 \\
\cmidrule{1-8}
0.8 & Baseline & 58.21 & 0.9996 & 0.376 & 174 & 335 & 509 \\
0.8 & + SSM \& EAR & 59.61 & 0.9997 & 0.365 & 188 & 358 & 546 \\
0.8 & MambaRaw & 59.58 & 0.9997 & 0.361 & 158 & 309 & 467 \\
\cmidrule{1-8}
5.0 & Baseline & 59.42 & 0.9997 & 0.705 & 191 & 367 & 558 \\
5.0 & + SSM \& EAR & 60.55 & 0.9998 & 0.685 & 205 & 392 & 597 \\
5.0 & MambaRaw & 60.52 & 0.9998 & 0.672 & 175 & 338 & 513 \\
\cmidrule{1-8}
20.0 & Baseline & 59.85 & 0.9998 & 1.052 & 205 & 394 & 599 \\
20.0 & + SSM \& EAR & 61.12 & 0.9999 & 0.985 & 221 & 420 & 641 \\
20.0 & MambaRaw & 61.08 & 0.9999 & 0.965 & 192 & 365 & 557 \\
\bottomrule
\end{tabular}
\end{table}

\begin{table}[!htbp]
\scriptsize
\centering
\setlength{\tabcolsep}{2.8pt}
\renewcommand{\arraystretch}{1.05}
\caption{Detailed ablation on Samsung NX2000 across all $\lambda$ values.}
\label{tab:ablation_samsung}
\begin{tabular}{cccccccc}
\toprule
\textbf{$\lambda$} & \textbf{Config} & \textbf{PSNR (dB)} & \textbf{SSIM} & \textbf{bpp} & \textbf{Enc. (ms)} & \textbf{Dec. (ms)} & \textbf{Total (ms)} \\
\midrule
0.02 & Baseline & 53.85 & 0.9981 & 0.188 & 215 & 402 & 617 \\
0.02 & + SSM \& EAR & 54.45 & 0.9986 & 0.182 & 232 & 428 & 660 \\
0.02 & MambaRaw & 54.42 & 0.9982 & 0.178 & 195 & 365 & 560 \\
\cmidrule{1-8}
0.8 & Baseline & 56.74 & 0.9996 & 0.376 & 228 & 424 & 652 \\
0.8 & + SSM \& EAR & 57.95 & 0.9997 & 0.372 & 245 & 452 & 697 \\
0.8 & MambaRaw & 57.91 & 0.9997 & 0.361 & 208 & 385 & 593 \\
\cmidrule{1-8}
5.0 & Baseline & 57.82 & 0.9997 & 0.725 & 248 & 465 & 713 \\
5.0 & + SSM \& EAR & 58.85 & 0.9998 & 0.705 & 265 & 495 & 760 \\
5.0 & MambaRaw & 58.82 & 0.9998 & 0.692 & 225 & 422 & 647 \\
\cmidrule{1-8}
20.0 & Baseline & 58.25 & 0.9998 & 1.125 & 268 & 507 & 775 \\
20.0 & + SSM \& EAR & 59.45 & 0.9999 & 1.085 & 285 & 538 & 823 \\
20.0 & MambaRaw & 59.41 & 0.9999 & 1.065 & 242 & 460 & 702 \\
\bottomrule
\end{tabular}
\end{table}

\begin{table}[!htbp]
\scriptsize
\centering
\setlength{\tabcolsep}{2.8pt}
\renewcommand{\arraystretch}{1.05}
\caption{Detailed ablation on Olympus E-PL6 across all $\lambda$ values.}
\label{tab:ablation_olympus}
\begin{tabular}{cccccccc}
\toprule
\textbf{$\lambda$} & \textbf{Config} & \textbf{PSNR (dB)} & \textbf{SSIM} & \textbf{bpp} & \textbf{Enc. (ms)} & \textbf{Dec. (ms)} & \textbf{Total (ms)} \\
\midrule
0.02 & Baseline & 56.15 & 0.9984 & 0.185 & 173 & 334 & 507 \\
0.02 & + SSM \& EAR & 57.25 & 0.9988 & 0.178 & 189 & 356 & 545 \\
0.02 & MambaRaw & 57.21 & 0.9985 & 0.175 & 162 & 315 & 477 \\
\cmidrule{1-8}
0.8 & Baseline & 59.04 & 0.9997 & 0.376 & 181 & 348 & 529 \\
0.8 & + SSM \& EAR & 60.52 & 0.9998 & 0.368 & 195 & 370 & 565 \\
0.8 & MambaRaw & 60.45 & 0.9998 & 0.361 & 168 & 325 & 493 \\
\cmidrule{1-8}
5.0 & Baseline & 60.18 & 0.9998 & 0.585 & 195 & 376 & 571 \\
5.0 & + SSM \& EAR & 61.35 & 0.9999 & 0.565 & 210 & 398 & 608 \\
5.0 & MambaRaw & 61.30 & 0.9999 & 0.552 & 180 & 350 & 530 \\
\cmidrule{1-8}
20.0 & Baseline & 60.65 & 0.9999 & 0.925 & 208 & 401 & 609 \\
20.0 & + SSM \& EAR & 62.15 & 0.9999 & 0.885 & 225 & 428 & 653 \\
20.0 & MambaRaw & 62.10 & 0.9999 & 0.865 & 192 & 375 & 567 \\
\bottomrule
\end{tabular}
\end{table}

\section{More Visualization Results}
\begin{figure}[H]
\centering
\includegraphics[width=1.0\linewidth]{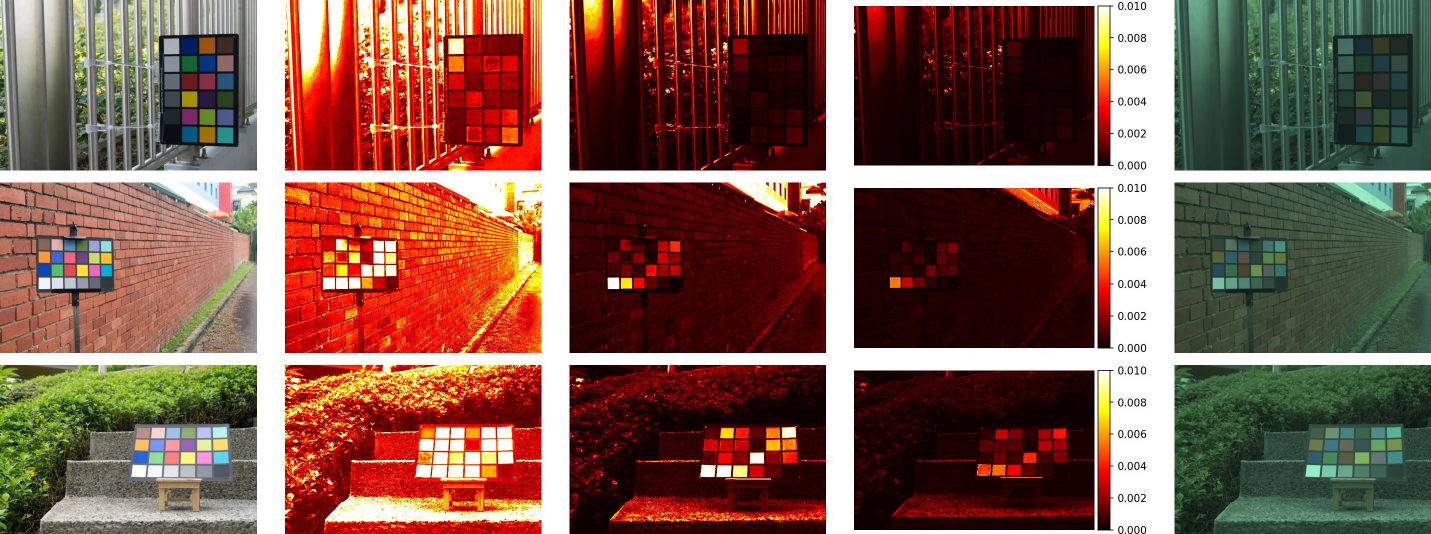}
\QualLegendRow
\caption{Additional qualitative results on Sony SLT-A57 in the same setting as Figure~\ref{fig:vis_sony}. Error maps visualize the per-pixel maximum absolute error over the three channels (after gamma correction for visibility); darker indicates smaller error. We keep the same error scale across methods to enable direct comparison.}
\label{fig:vis_sony_appendix}
\end{figure}

We provide additional qualitative comparisons to further demonstrate the reconstruction fidelity of our proposed MambaRaw framework. Figure~\ref{fig:vis_sony_appendix} presents more visual examples from the Sony SLT-A57 subset. Consistent with the observations in the main text, our method effectively preserves high-frequency details and complex textures, resulting in visibly lower residual errors compared to the baseline methods. This further validates that the spatial-energy coupled context modeling in MambaRaw can robustly capture fine structural information across diverse scenes.

\end{document}